\documentclass[journal]{IEEEtran}
\usepackage{cite}
\usepackage[pdftex]{graphicx}
\usepackage[cmex10]{amsmath}
\usepackage{array}
\usepackage{fixltx2e}
\usepackage{color}
\usepackage{url}

\begin{document}
\title{Ethnicity and Biometric Uniqueness:  \ Iris Pattern Individuality
in a West African Database}
\author{John Daugman, \ Cathryn Downing, \ Oluwatobi Noah Akande, \ Oluwakemi Christiana Abikoye%

\thanks{Submitted to {\em IEEE Transactions on Biometrics, Behavior, and Identity Science (T-BIOM)}
on January 18, 2023.  Revised: March 29, 2023.  Revised again:  April 30, 2023.  Revised again: July 6, 2023.
All but one of the reviewers have rated the manuscript as ``Excellent" and have recommended:  ``Accept with
No Changes."  One isolated reviewer raises constantly changing objections, and now is unresponsive.  So,
eight months after the original submission, with no decision yet from {\em T-BIOM}, we have decided to
release this paper on ArXiv.  We note that other authors have taken the same path.} }

\IEEEpubid{0000--0000/00 }

\markboth{Ethnicity and Biometric Uniqueness: \  Iris Pattern Individuality in a West African Database}{}
\maketitle
\begin{abstract}
We conducted more than 1.3 million comparisons of iris patterns encoded 
from images collected at two Nigerian universities, which constitute
the newly available African Human Iris (AFHIRIS) database.  The purpose
was to discover whether ethnic differences in iris structure and appearance
such as the textural feature size, as contrasted with an all-Chinese image
database or an American database in which only 1.53\% were of African-American
heritage, made a material difference for iris discrimination.  We measured
a reduction in entropy for the AFHIRIS database due to the coarser iris features
created by the thick anterior layer of melanocytes, and we found stochastic
parameters that accurately model the relevant empirical distributions. 
Quantile-Quantile analysis revealed that a very small change in operational decision
thresholds for the African database would compensate for the reduced entropy and
generate the same performance in terms of resistance to False Matches.  We conclude
that despite demographic difference, individuality can be robustly discerned
by comparison of iris patterns in this West African population. 
\end{abstract}

\section{Introduction}

\IEEEPARstart{A}{}question of increasing salience today for deployment of biometric identification technologies
relates to ethnicity;  specifically, whether False Match probabilities are worse for some ethnic groups
than for others.   This has become a particularly notorious problem for face recognition algorithms,
with several reports \cite{Leslie} \cite{Castel} \cite{Cavazos} that persons of African descent are
much more likely than others to be mis-identified, or even to be mis-classified by image classification
systems as gorillas \cite{BBC}.  A more subtle problem is bias among system designers, or data
bias creating limited representation in the image training datasets, particularly given the dominance
today of machine learning approaches.  If a training set is unrepresentative of some ethnic groups
(or indeed of gender, age, etc), then performance is demonstrably worse for the poorly represented
groups \cite{Monea}.   This problem is compounded by the ``black box" nature of deep machine learning
methods:   it is difficult to know, or even to try to discover, what such algorithms have actually
extracted from their training datasets.

These problems clearly raise public policy issues, and can generate scepticism, suspicion, or even
resistance against biometric deployments.  There are also purely technical questions related to ethnicity
in biometrics, because some traits vary in visibility among different ethnic groups, with possible effects
on identifiability.  An obvious challenge for iris recognition, for example, is that persons of East Asian
heritage frequently have much eyelid occlusion, which may leave less than 50\% of the iris visible.
Likewise, persons of sub-Saharan African or of Malaysian heritage usually have dense melanin pigmentation,
which limits the visibility of iris texture in visible wavelengths.   Most iris cameras use near-infrared
illumination wavelengths (NIR:  700 – 900nm) in which melanin is almost completely non-absorbing, and
therefore less problematic.  Nonetheless the anterior layer of the iris in persons of such descents contains
a thick blanket of chromatophore cells (superficial melanocytes) \cite{Snell} that create a coarser texture
of crypts and craters as seen in Fig.~1, almost lunar in appearance, rather than the fine fibrous details
more typically visible in the iris of persons having (say) “blue eyes” which lack such a thick anterior layer.
\IEEEpubidadjcol
\begin{figure*}[t!]
\centering 
\includegraphics[scale=0.86]{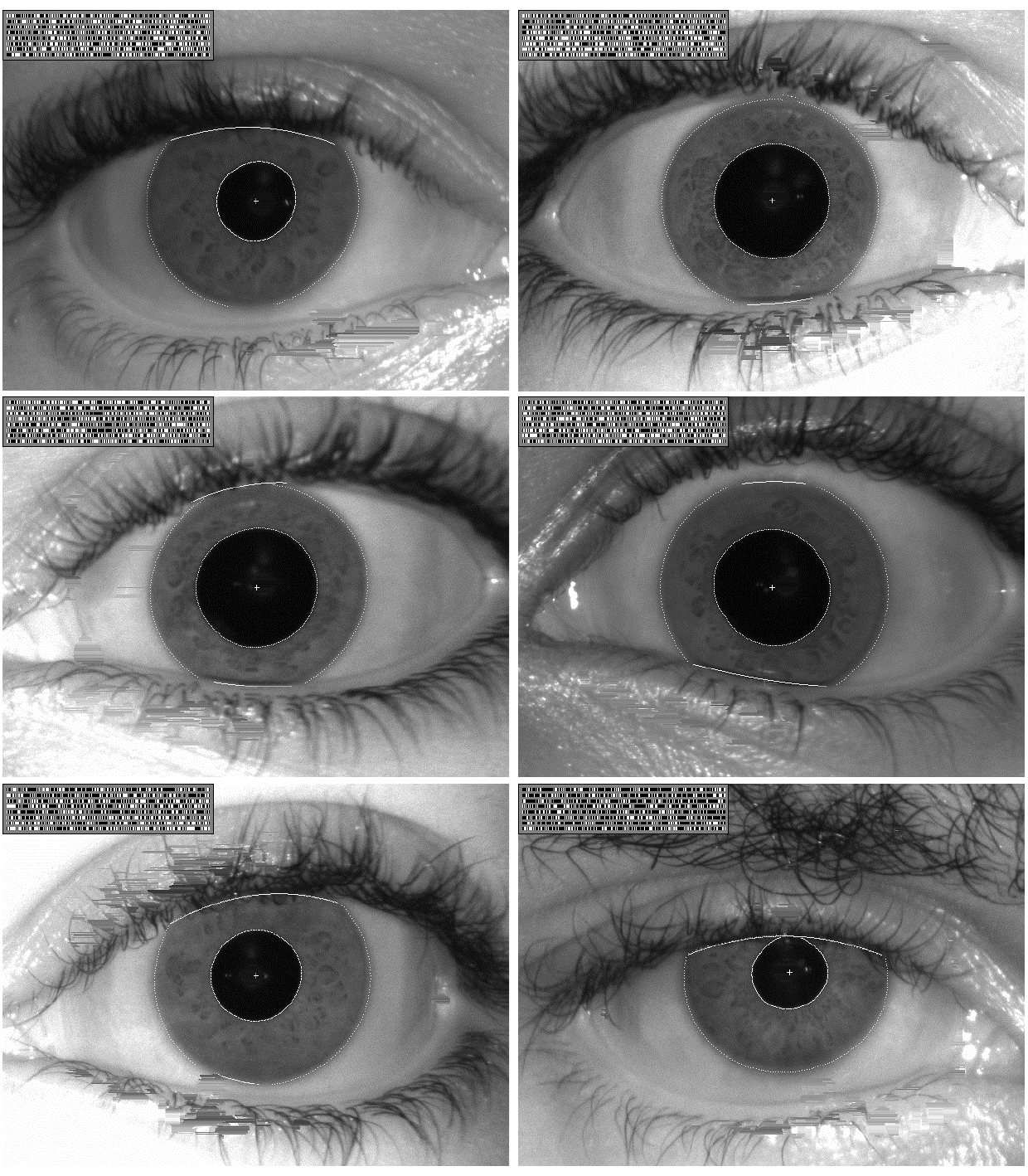}
\caption{Sample images in the AFHIRIS dataset, illustrating the coarse ``lunar" texture of craters and crypts.
White contours show the results of automatic image segmentation.  IrisCode bit streams are depicted as binary
pixel sequences in the upper-left of each image.}
\end{figure*}

There are core mathematical and scientific issues associated with ethnic differences in the traits encoded
by biometric technologies.   The random variation that is the basis of all biometric discriminability among
persons may vary in its complexity or dimensionality across different ethnic groups.  It is even conceivable
that within some unstudied demographic group, all persons might have identical or very similar iris patterns.
Face recognition is bedevilled by genetic determinism of facial appearance at a given age, causing current
algorithms to fail utterly to distinguish between monozygotic (MZ) twins, and even between most dizygotic
twins \cite{Grother}.  More than 2,000 genes are expressed in the iris \cite{SturmLarsson}, and it is
well-known that a person’s biogeographic ancestry is correlated with iris colour and general textural
appearance \cite{Quillen}.  It is possible that strong genetic coherence in some ethnic or social groups 
which mainly inbreed (e.g. Quilombo;  Amish; Haredi) \cite{Inbreeding} \cite{Lemes} might limit variation
among their iris patterns.  Although iris pattern detail seems to be epigenetic, as evidenced by the
observation that even genetically identical eyes such as those of MZ twins or the two possessed by one
person are mathematically uncorrelated in their detailed iris texture \cite{DaugDown2001} \cite{DaugDown2020},
iris individuality has not been widely investigated within different demographic groups.

Tools for the quantitative analysis of random variation among patterns, and for comparison of their relative
complexity (specifically entropy) of random variation, are provided by Information Theory \cite{CovThom} 
\cite{Daug2015}.  The purpose of this paper is to apply such tools to a newly available database of iris
images acquired from persons of African descent (Nigerian university students), a demographic group
previously unstudied in such biometric research.  Our principal question is whether the entropy among their
iris patterns differs significantly from that of other demographic datasets in which persons of African
descent are absent, such as the Chinese iris image datasets collected by the Chinese Academy of Sciences'
Institute of Automation (CASIA) \cite{CASIA} \cite{CASIA-dataset}, or the large University of Notre Dame 
database \cite{BowyerFlynn} \cite{Connaughton} of iris images in which the ethnic identity
of all subjects are tagged but only 1.53\% are tagged as of ``black or African-American" descent.
We hope that this paper contributes to understanding whether, and to what extent, ethnicity affects 
the ability of iris recognition to discriminate individuals or allows False Matches instead.

\section{Databases and Methods}

The African Human Iris (AFHIRIS) database \cite{AFHIRIS} is the first of its kind, and it was made freely
available in 2022.  It was collected from 1,028 student and staff volunteers (58\% male, 42\% female) at
Ladoke Akintola University of Technology, Oyo State, Nigeria, and at Landmark University in Omu-Aran, Kwara
State, Nigeria.   About half of the subjects were aged under 21 years, with the remainder aged 21 - 45 years.
They originated from 34 of the 36 States in Nigeria.  

Images were captured using a Corvus VistaEY2H handheld dual iris camera.  In half of these the subject was
wearing spectacles, but because of the camera’s frontal illumination system, very few of these were useable
since the iris was heavily obscured by specular reflections from the eyeglasses.   Among the other half of the
images, acquired without spectacles, 20\% were in very poor focus and were discarded.  Additionally, eleven
persons had been enrolled under multiple identities by the student enrollers.  After confirmation, those
images were also discarded as ground-truth errors.   There remained a total of 1,648 images (each one of
a different eye) deemed to be of sufficient quality for this study.   Samples illustrating the best image
quality are provided in Fig.~1.    Images were automatically processed and enrolled into a database of
IrisCodes using the classical methods that have been described previously \cite{Daug2003} \cite{Daug2004}
\cite{Daug2007} and which are used worldwide in all publicly deployed systems for iris recognition. 
``All-against-all” cross-comparisons were then performed on all possible pairings among these 1,648 images,
making a total of 1,357,128 unique pairings, whose Hamming distances \cite{Daug2004} were tabulated and
plotted as a histogram in the usual manner.  Included among them were comparisons between the right and left
eyes of individual persons, but these amounted to only 0.06\%  (i.e. 824 / 1,357,128) of all the
IrisCode pairings.  One photograph, binocular and without spectacles, was taken for each person so it is
not possible to measure any same-eye Hamming distances. 

We also used a Notre Dame database \cite{BowyerFlynn} \cite{Connaughton} (29,986 images of 1,352 different
eyes) and a CASIA database \cite{CASIA} \cite{CASIA-dataset} (3,183 images of 400 different Chinese eyes)
to contrast biometric entropies, hence discriminability among different persons, compared with AFHIRIS.
Those image databases were also without spectacles.  Their gender and age distributions were similar to
AFHIRIS, being comprised mainly of university students.  The same classical IrisCode algorithm
\cite{Daug2003} \cite{Daug2004} \cite{Daug2007} was used for all of this work.  Recently there has
been great interest in alternative, automatically learned (not human-designed) Deep Learning (DL)
processes, which did lead to revolutionary improvements in face recognition and in some other fields
within computer vision; but they have not yet done so for iris recognition apart from some benefits
in segmentation and spoof detection.  Moreover, it is impossible to see into such ``black boxes"
to understand what they have learned from their training data, encoded into millions of learned
parameters, across sometimes hundreds of layers.  Some independent researchers \cite{IEG} have described
this current situation as the dichotomy between ``Deep Learning and deep understanding".

Quoting from a recent very comprehensive survey \cite{DeepLearning} of more than 200 papers 
applying DL to iris recognition:  Most of the DL methods ``do not work under
the one-shot learning paradigm;  assume multiple observations of each [eye] to obtain
appropriate decision boundaries;  and -- most importantly -- have encoding / matching steps
with time complexity that forbid their use in large environments (in particular, for
all-against-all settings)" \cite{DeepLearning}.  Additionally, tests by the US National Institute
of Standards and Technology (NIST) \cite{IREX-10} indicate that the DL-based submissions
(identifiable by their heavyweight model size with millions of learned parameters) fail to
distinguish between genetically identical eyes.   Unlike the classical IrisCode algorithm, they
return similarity scores that are much closer for twins' eyes, and also when comparing the two eyes
of a given person, than for unrelated eyes.  Probably this is because the DL ``black boxes" are
encoding and matching periocular data from eye images, such as the shape of the eyelids.  

Finally, and of central importance for the topic of this study, the DL methods perform poorly on
NIST's new ethnic ``Equity Measure for False Positives".  For a given algorithm, this measure is
defined as the factor by which the worst False Match Rate (FMR) suffered by any ethnic group is
worse than the geometric mean FMR across all groups using that algorithm.  Obviously a value
near 1.0 represents ethnic fairness, while larger values progressively signify inequity.  The NIST
report's Demographics section \cite{IREX-10} shows that DL methods often produce inequity factors
of 2.0 or higher, whereas the classical IrisCode remains closer to 1.0 which signifies equal resistance
to False Matches across ethnic groups.  Given the focus of the present paper, obviously we avoided
DL methods.  The classical and widely deployed IrisCode approach enables baseline comparisons of entropy
estimates, which in turn reveal likelihoods that two different biometric identities may collide
by chance.  The findings which we will present here, showing only a small impact of ethnicity on iris
discriminability, align with the new NIST Equity Measure for False Positives \cite{IREX-10}
testing the same core IrisCode algorithm.

\section{Results}

The histogram of Hamming distances (HD) across all 1,357,128 possible pairwise comparisons
of IrisCodes in the AFHIRIS database is plotted in Fig.~2.   As expected, its mean is close
to 0.5 because the data bits in IrisCodes are equally likely to be a 0 or 1, and thus when 
corresponding bits from any two independent IrisCodes are compared, their four joint possibilities
(00, 01, 10, 11) are all equiprobable, so therefore half of such paired bits are expected to
disagree (HD = 0.5).  Indeed comparing pairs of bits using their Exclusive-OR to detect disagreement, 
which reduces the above four pairs to simply (0, 1, 1, 0) amounts itself to a toss of a fair coin. 
In Fig.~2 it is clearly very unlikely that fewer than 40\% of the bits or more than 60\% of
the bits compared from independent IrisCodes (HD $<$ 0.4 or HD $>$ 0.6) will disagree by chance.  
This is for exactly the same reason that if one tosses a ``fair coin” (meaning its probability of
coming up Heads is $p=0.5$)  enough times, the outcomes are very unlikely to deviate far from a 50\%
frequency of Heads. The critical consequence for biometric iris recognition is that when two different 
IrisCodes are compared, if their Hamming distance is smaller than (say about) 0.3, then it is extremely 
unlikely that they arise from different eyes.   The probability that a sample from the distribution 
in Fig.~2 will be HD $<$ 0.3 is infinitesimally small.   This is the reason why iris recognition
technology (using these algorithms) has such a legendary resistance to False Matches \cite{Daug2015}
\cite{Daug2003} \cite{IREX-III} and can survive very large database searches without making any, 
for example in de-duplication operations by performing all-against-all cross-comparisons.
\begin{figure}
\centering
\includegraphics[scale=0.42]{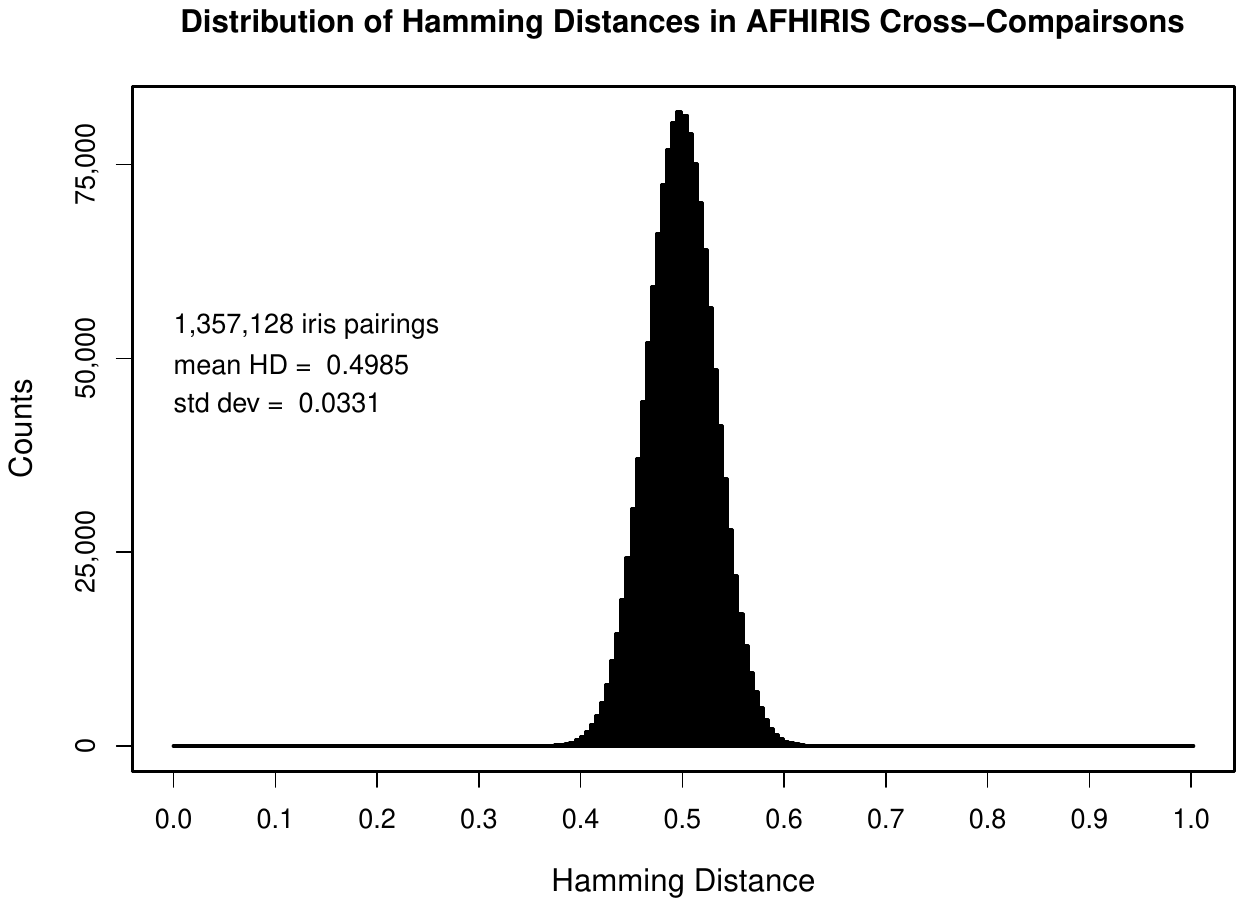}
\caption{Histogram of Hamming distances generated by all 1,357,128 pairwise comparisons
of IrisCodes computed from the AFHIRIS database.}
\end{figure}

Most informative is the distribution’s standard deviation, $\sigma = 0.0331$, because this reveals the
entropy within this database.   The greater the number of coin tosses, the narrower the distribution of 
fractional outcomes will be (regardless of whether the coin is ``fair”, or not).   In fact the std dev 
$\sigma$ varies inversely with the number of independent tosses.   This allows us to estimate the entropy 
of iris patterns (a measure of their amount of random variation) as the equivalent number of coin tosses
that would generate such a distribution when their IrisCodes are compared to compute a Hamming distance.
For the AFHIRIS database, it corresponds to 228 tosses of a fair coin in a run.   In terms of
Information Theory, that means 228 bits of entropy:   each toss of a fair coin has 1 bit of entropy
(but less if $p \ne 0.5$).   In Fig.~3, the same histogram as plotted in Fig.~2 is superimposed with
a curve showing the binomial probability density distribution corresponding to tossing a fair coin 228
times in each of many runs, and then tabulating the frequency of observing any given fraction (HD)
of Heads, after performing many such runs.

\begin{figure}
\centering
\includegraphics[scale=0.42]{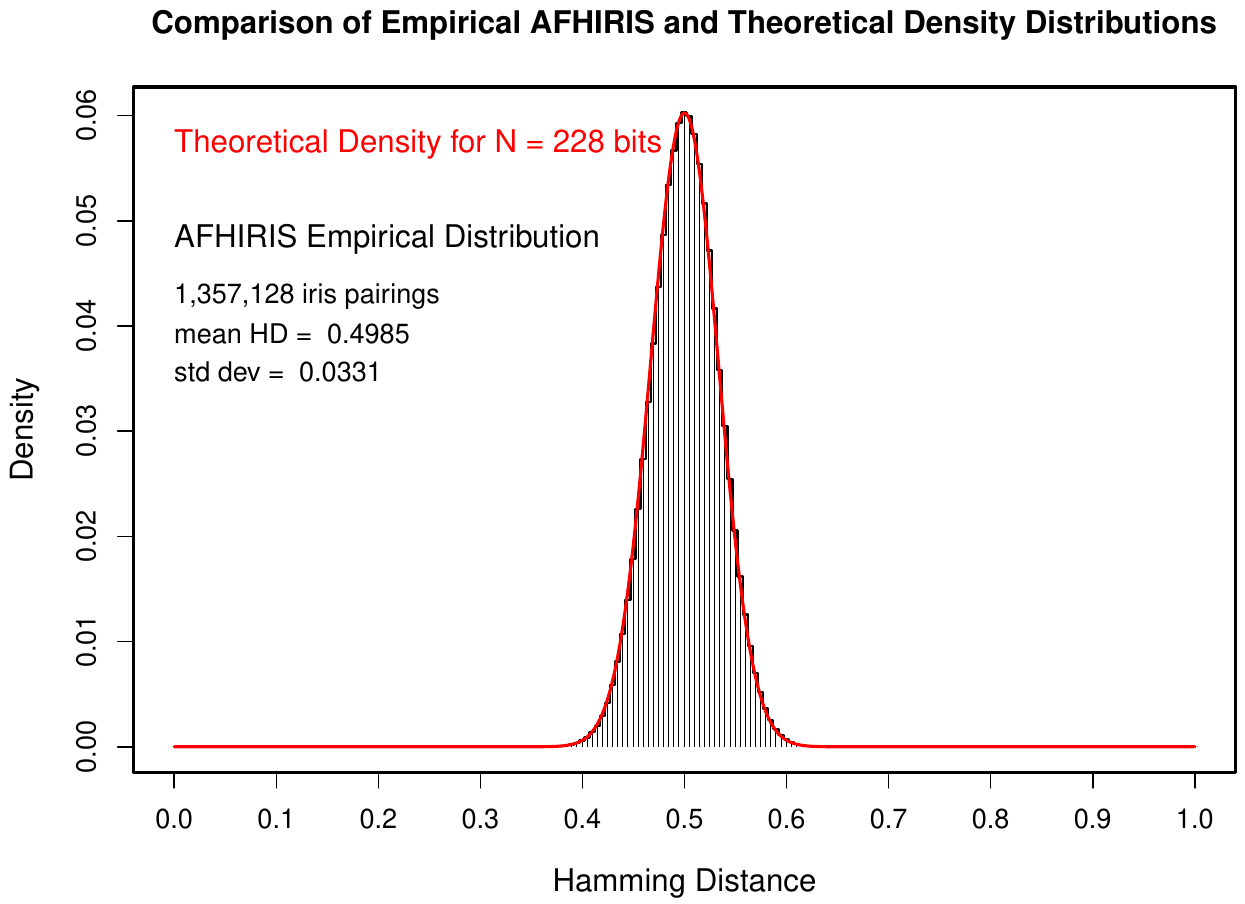}
\caption{The histogram from Fig.~2 combined with a theoretical bionomial probability density
distribution (red curve), which plots (1) using parameters $p=0.5$ and $N=228$.}
\end{figure}

The red curve in Fig.~3 is a plot of the following probability distribution $\mathrm{prob(HD)}$
for the fraction of Heads (HD) in a run of $N$ tosses of a coin whose probability of Heads is $p$\,:
\begin{equation}
\mathrm{prob(HD)} =\frac{N!}{m!(N-m)!}\hspace*{0.08in}p^{m}(1-p)^{(N-m)}
\end{equation}
where in this case $N=228,$ $p=0.5,$ and $\mathrm{HD} = m/N$ is the outcome fraction of $N$ Bernoulli
trials (e.g. observing $m$ Heads within a run of $N$ coin tosses).  (1) can be understood as the product of
a combinatorial term and a probability term.   The combinatorial term is simply the binomial coefficient 
for the number of different ways in which $m$ items can be selected out of $N$:
 \[ \left( {\begin{array}{c}
	 N\\
	 m\\
           \end{array} } \right) = \ \frac{N!}{m!(N-m)!} \]
This is multiplied by the  probability term, which expresses the joint probability that the outcome whose
probability is $p$ occured $m$ times while the alternative outcome whose probability is $(1-p)$ occured
the remaining $(N-m)$ times.  Thus the joint probability associated with any outcome sequence tabulated in 
the combinatorial term is $p^{m}(1-p)^{(N-m)}$. It is noteworthy that in the case that $p=0.5$, this joint
probability term does not depend on $m$, hence Hamming distance, at all.  The shape of the overall probability
distribution then arises entirely from the combinatorial term.  Measuring the std dev $\sigma$ for an
empirical distribution of HD scores from independent pairings tells us the equivalent number of
tosses of a coin (having probability $p$ of Heads), namely $N = p(1-p)/\sigma^{2}$.   In Fig.~2 we
measured $\sigma = 0.0331$, and therefore we estimated that the biometric entropy shown in the AFHIRIS
database by cross-comparisons among all pairings was $N = 228$ bits.  The fit in Fig.~3 between the empirical
distribution data and the theoretical probability density curve seems to be excellent.

The estimated entropy in a University of Notre Dame database of iris images (in which only 1.53\% were
tagged as of African descent) is larger, corresponding to $N = 260$ bits.  Its histogram of cross-comparison 
HD scores is shown in Fig.~4 (red), superimposed on that for AFHIRIS (black).  This difference may
reflect the coarser structure in African iris patterns, as seen in Fig.~1, because a fundamental link
exists in Information Theory between bandwidth and channel capacity.  Lower frequency (coarser) structure
is ``less busy," with slower variation, hence lower entropy.  But as we shall assess quantitatively, this
level of demographic difference in entropy has only a small operational impact on the deployment parameters
(decision thresholds) necessary to maintain the same powers of iris discrimination. 
\begin{figure}
\centering
\includegraphics[scale=0.42]{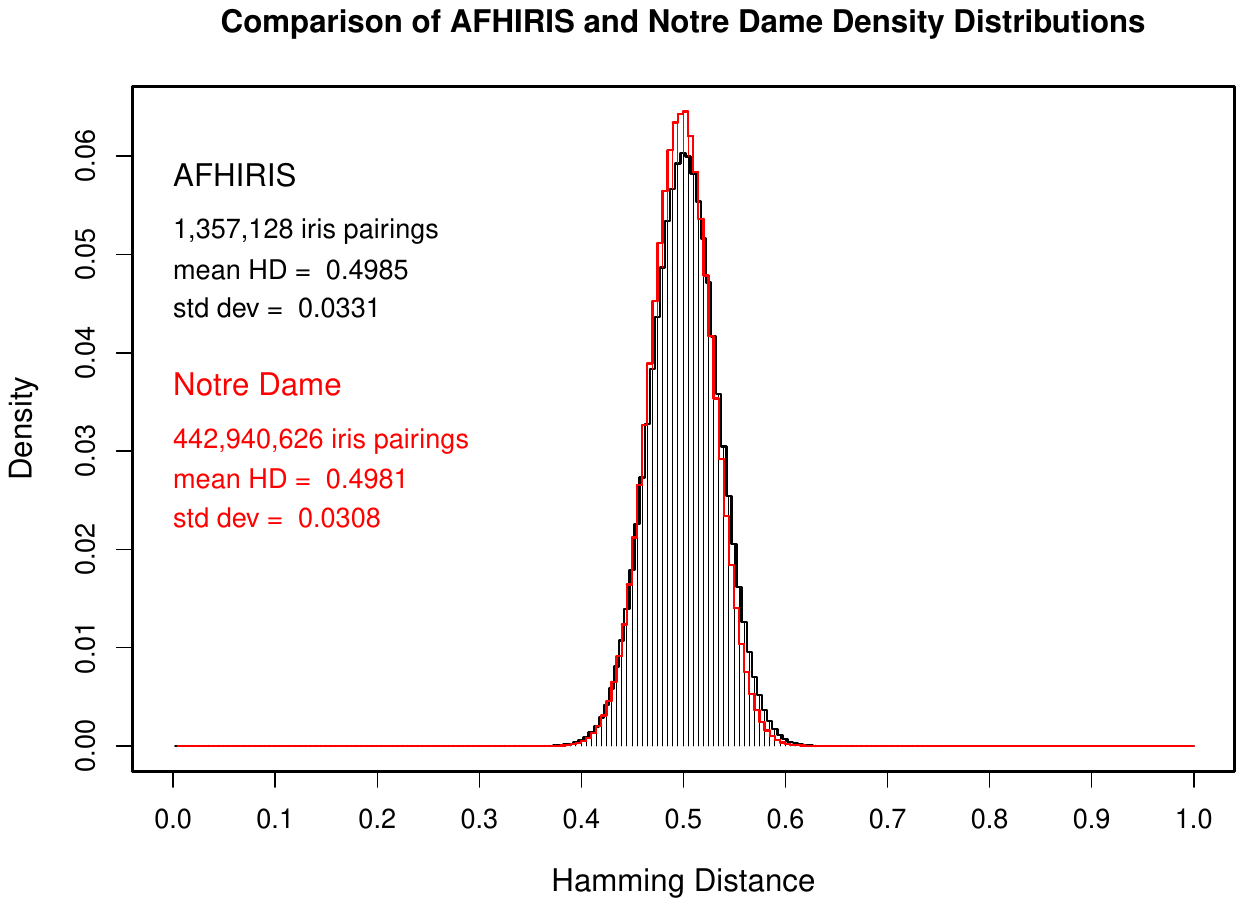}
\caption{The histogram from Fig.~2 for the AFHIRIS database (black bars), superimposed with 
the histogram from cross-comparisons \cite{DaugDown2020} within a large Notre Dame database 
\cite{BowyerFlynn} \cite{Connaughton} (red bars).}
\end{figure}

In Fig.~5 we also superimpose the distribution of AFHIRIS all-against-all cross-comparison scores
(black) with the distribution of such scores obtained from cross-comparisons (blue) within a database 
of all-Chinese eyes, acquired by the Chinese Academy of Sciences (CASIA) \cite{CASIA} \cite{CASIA-dataset}.
The contrast between these two distributions is similar to that observed in Fig.~4.  The AFHIRIS distribution 
reveals a somewhat smaller entropy:  a somewhat larger stn dev $\sigma$.  (The small shifts in the mean HD
score below 0.5 for all three of these distributions seem to arise from illumination gradients created 
by the local acquisition conditions.  Shared gradients within an image dataset reduce cross-comparison
HD scores slightly, because the wavelet encodings are sensitive to spatial derivatives even on a gross
scale \cite{Daug2004}.) The main contrast observable in Fig.~5, the larger stn dev $\sigma$ for the
AFHIRIS database, can be related to the coarser scale African iris texture features as illustrated
in Fig.~1 relative to those in Chinese eyes as illustrated in the samples provided at \cite{CASIA-dataset}.
\begin{figure}
\centering
\includegraphics[scale=0.42]{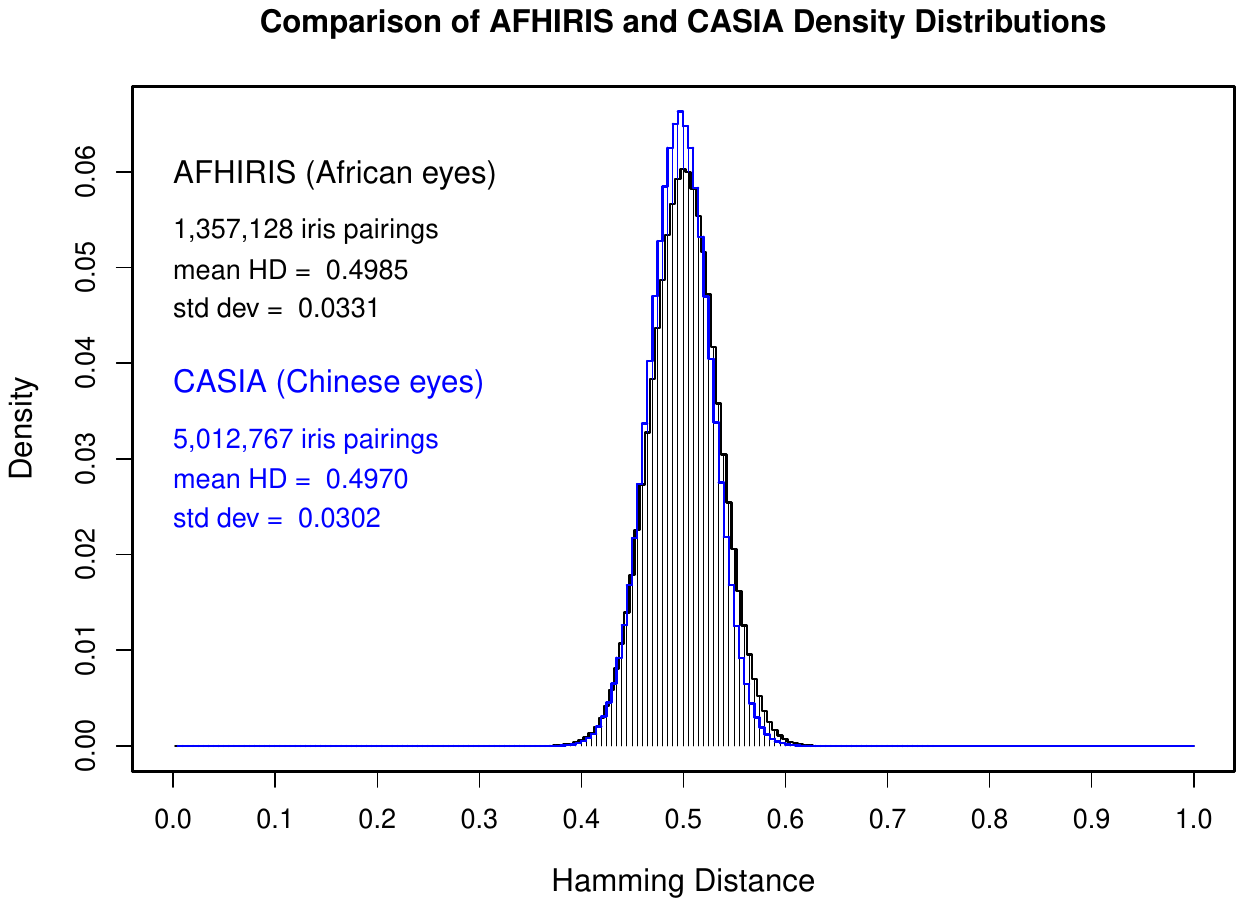}
\caption{The histogram from Fig.~2 for the AFHIRIS database (black bars), superimposed with the
histogram of cross-comparisons within a database of all-Chinese eyes \cite{CASIA} \cite{CASIA-dataset}
(blue bars).}
\end{figure}

\section{Discussion}

Another aspect of some Nigerian iris patterns which may contribute to some loss of entropy relative to
Caucasian or Chinese eyes eyes is the occasional appearance of broad radial ``stripey" patterns as
illustrated in Fig.~6.  Just as with the coarse pattern of ``lunar" craters seen in Fig.~1, their
large scale reduces entropy by imposing a slower form of variation.   To the extent that such patterns
confine variation to just a single (polar) variable, namely variation along the angular coordinate
but much less along the radial coordinate, such structures reduce information capacity for the
same reason that a 1-D bar code is a channel with less information capacity than a 2-D bar code.
\begin{figure*}
\centering
\includegraphics[scale=0.70]{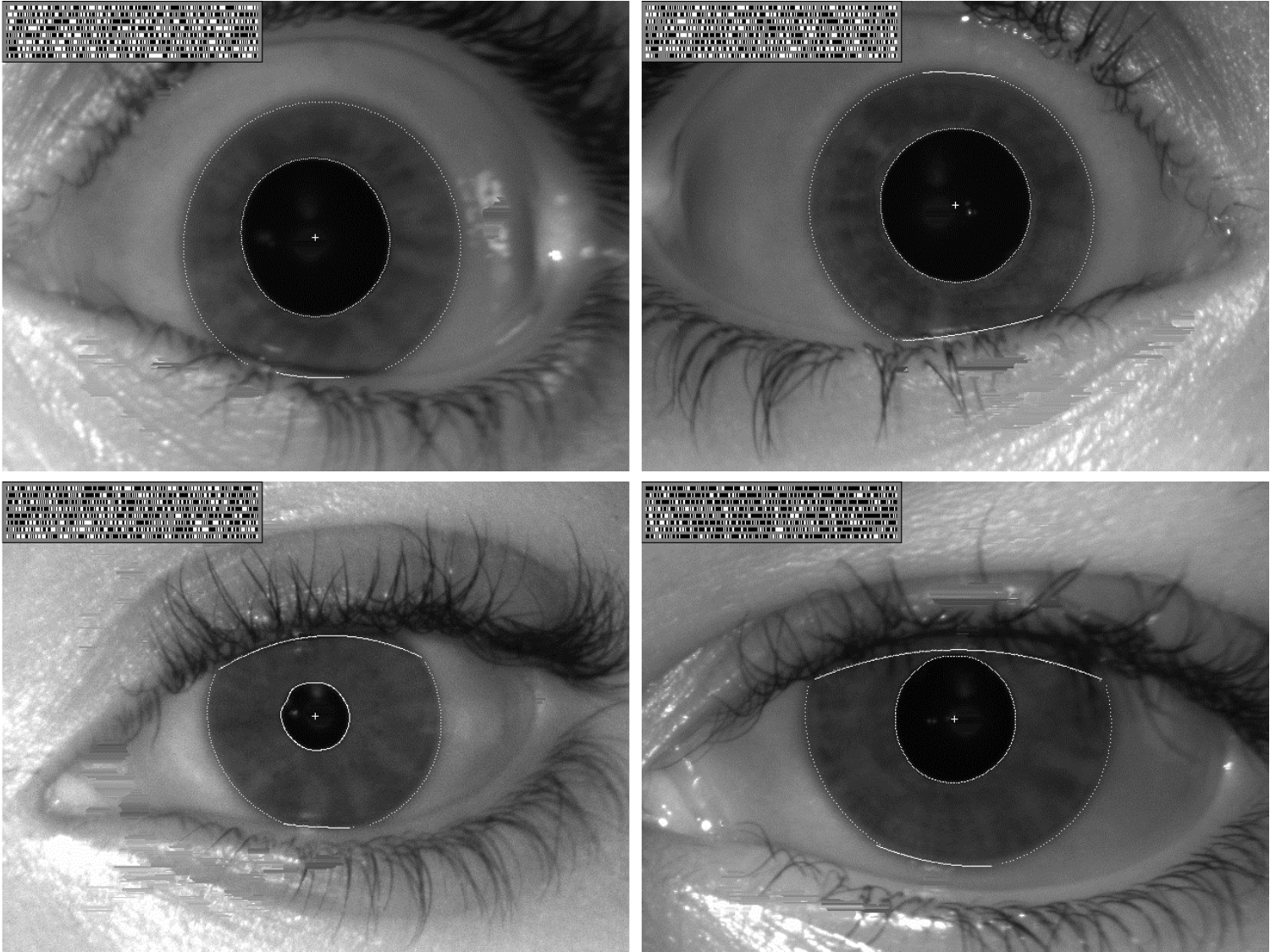}
\caption{Sample images in the AFHIRIS dataset illustrating coarse ``stripey" radial structure.
White contours show the results of automatic image segmentation.}
\end{figure*}

IrisCodes must be compared over a range of possible ``tilt angles" (polar rotations) for a match, 
because the camera or the head may be tilted, and indeed each eye has a range of \emph{cyclotorsion}
(rotation around its optical axis), reflecting the fact that each eyeball's pose within its orbital
socket is controlled by not just two but three pairs of attached muscles \cite{Snell}. 
Therefore the search engine attempts multiple matches after scrolling each IrisCode over some
reasonable range of relative rotations, in steps of $360^{\mathrm{o}}/128 \approx 2.81$ degrees.
Usually seven rotations are attempted before assessing the ``best match" possible between any two
IrisCodes, but more may be required with handheld monocular cameras (as used in common smartphones)
because they are often held at bigger tilt angles.   Retrieving the best match after seven comparison
rotations spans a tilt range of about 20 degrees. 

Attempting multiple matches over some range of candidate rotations and recording only the closest match
amounts to \emph{Extreme Value Sampling} of the histogram distributions shown earlier.  Obviously
the mean HD score is reduced and a skew bias is introduced, as can be seen in the histogram of Fig.~7,
because only the smallest sample from each set of sample scores is retained.

The analysis of this effect can be given a general theoretical form.  Let $f_{1}(x)$ be whatever
density distribution is obtained for HD scores $x$ between different IrisCodes when compared for
only a single relative image tilt.  For example, $f_{1}(x)$ might be the fractional binomial defined as
prob(HD) in (1).  Its cumulative $F_{1}(x)$ is the probability of getting a score of $x$ or smaller:
\begin{equation}
F_{1}(x)=\int_{0}^{x}f_{1}(x) dx
\end{equation}
or, equivalently,
\begin{equation}
f_{1}(x)=\frac{d}{dx}F_{1}(x) \ .
\end{equation}

The probability of \emph{not} getting a score that is smaller than $x$ is therefore
$1-F_{1}(x)$ in single comparisons, and it is $\left[1-F_{1}(x)\right]^{k}$
after carrying out $k$ such tests independently when considering $k$ different
relative tilt angles.  Thus the cumulative probability distribution $F_{k}(x)$ for
observing an HD score that is $x$ or smaller after optimising for relative image tilt is
\begin{equation}
F_{k}(x)=1 - \left[1 - F_{1}(x)\right]^{k} ,
\end{equation}
and the probability density distribution $f_{k}(x)$ expected for this cumulative is:
\begin{eqnarray}
f_{k}(x) & = & \frac{d}{dx}F_{k}(x)  \nonumber \\
 & = &  kf_{1}(x)\left[1 - F_{1}(x)\right]^{k-1} .
\end{eqnarray}

In the specific case that the raw density distribution $f_{1}(x)$ for HD scores is the fractional
binomial density prob(HD) as defined in (1), with parameter $N = 228$, and $k = 7$ relative
tilt angles before selecting the best match in each such group of tilts, the predicted Extreme
Value probability distribution $f_{k}(x)$ is the red curve plotted in Fig.~7.  It seems to provide 
an excellent fit to the empirical data (black bars) for such matches after rotations, and it
supports further conclusions.
\begin{figure}
\centering
\includegraphics[scale=0.42]{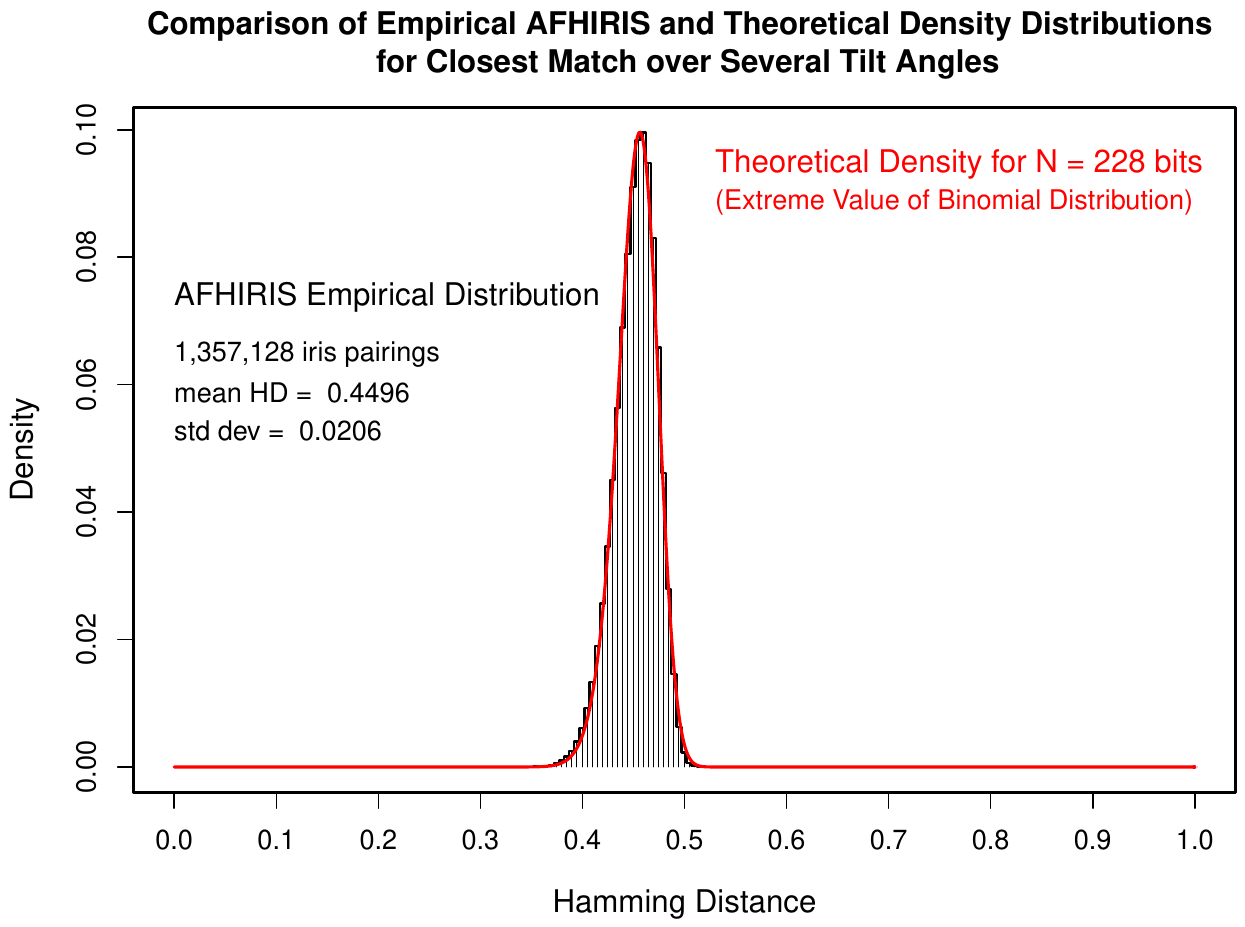}
\caption{Histogram of best HD scores obtained after attempting seven relative image tilts.
The red curve is (5), the theoretical Extreme Value density $f_{k}(x)$ associated with $f_{1}(x)$
as defined in (1).}
\end{figure}

\section{Conclusions}

Considering biometric identification in the context of Information Theory, iris patterns can be regarded
as \emph{communication channels} \cite{Daug2015} where identity is the ``signal" being transmitted, and
whose \emph{channel capacity} to distinguish different signals is measured in bits (or bits/mm$^{2}$). 
If channel entropy estimated in bits (the equivalent number of Bernoulli trials in Fig.~3) is larger, then 
more signals can be reliably discriminated.  The distribution of cross-comparison HD scores becomes narrower
(reduction in $\sigma$), so its cumulative up to any given HD decision criterion is reduced, both in the
raw (Fig.~3) and post-rotations (Fig.~7) distributions, reducing probability of False Matches.

A Kolmogorov-Smirnov test assessing the likelihood that the two collections of samples shown in Fig.~4 (one from
cross-comparisons within the Notre Dame database, and the other from within AFHIRIS) could be drawn from the same
distribution function, rejects this null hypothesis at astronomical levels.  This is not surprising, given the
large number of cross-comparison scores in each distribution.  The same conclusion arises from performing a
Kolmogorov-Smirnov test to compare the CASIA and AFHIRIS distributions shown in Fig.~5.  Regarding operational
significance of the AFHIRIS entropy reduction, it is useful to construct Quantile-Quantile (QQ) plots as presented
in Fig.~8 for Notre Dame and AFHIRIS.  They plot the HD decision criterion which, if adopted for one distribution,
would reach the same cumulative value (hence the same probability of False Matches) as the other distribution
would reach at some other criterion.  Fig.~8 shows this both for the raw distributions (left panel) and for
the post-rotations distributions (right panel).

Choosing a particular HD value may provide an illustrative example.  Measuring offsets in the post-rotation
QQ plot of Fig.~8 shows that to achieve False Match resistance comparable to that obtained when
operating at a decision threshold of around (say) HD~=~0.39 for the Notre Dame database (which contained
overwhelmingly Caucasian eyes, and only 1.53\% African-American eyes), the decision threshold should just
be reduced to around HD~=~0.38 for eyes in this West African demographic group.  We conclude that with such
minor adjustments to compensate for reduced entropy, associated with the comparative scale of features, iris
recognition technology can be deployed in this demographic group without any necessary compromise to its
now legendary resistance to False Matches.  This conclusion aligns with the recent NIST presentation
\cite{IREX-10} of its ethnic Equity Measure for False Positives, which hovers near 1.1 for the IrisCode
algorithm.  But this inequity factor ranges up to 2.0 or 3.0 for DL-based submissions.
Although the present AFHIRIS database is too small to test False Match rates at the 1-in-a-million
level (HD $< 0.33$) or 1-in-a-billion level (HD $< 0.30$), it is noteworthy that NIST did validate and confirm
\cite{Daug2015} those rates as were predicted long ago \cite{Daug2003} by the methods discussed here, when
NIST completed 1.2~trillion iris comparisons \cite{IREX-III} across geographically disjoint populations.
\begin{figure}
\centering
\includegraphics[scale=0.40]{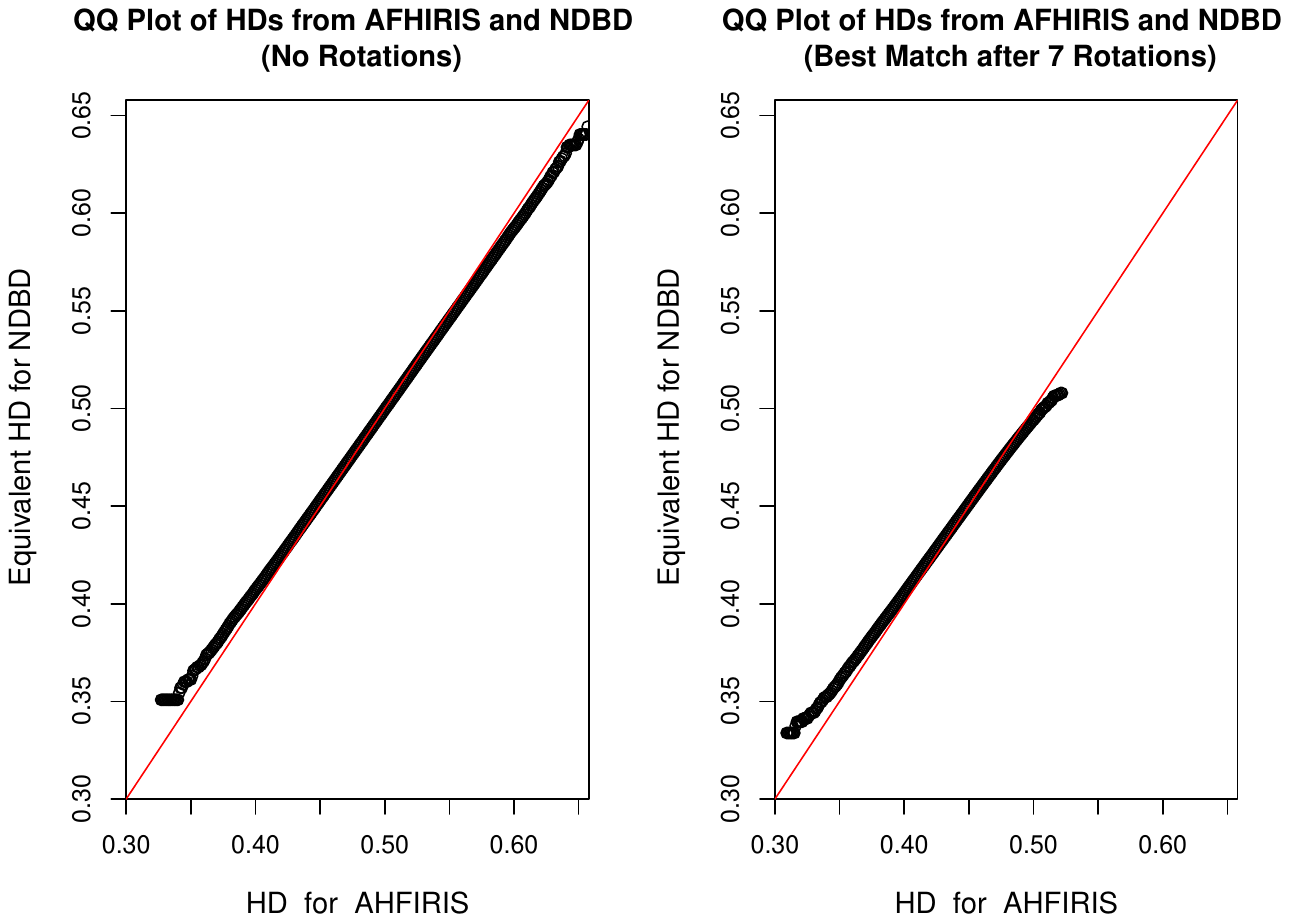}
\caption{Quantile-Quantile plots for the AFHIRIS and Notre Dame (NDBD) datasets, revealing which HD
decision criterion for one dataset would yield the same probability of False Matches as would some HD
criterion for the other dataset.} 
\end{figure}

\section*{Acknowledgements}

We are grateful to Dr.\ Emine Krichen (of Idemia) for valuable comments about comparative scores from
the NIST Equity Measure \cite{IREX-10}; and to the Chinese Academy of Sciences Institute of Automation,
and to the University of Notre Dame, for making biometrics databases available \cite{CASIA} 
\cite{CASIA-dataset} \cite{BowyerFlynn} \cite{Connaughton}.

Statistical analyses, and the generation of figures, were performed using the ‘R’ 
package: \ \url{https://cran.r-project.org/}

Institutional Review Board approvals: 
Ladoke Akintola University of Technology (ref: LODLC/ERC/2022/024), and Landmark University
(ref: LMUERC/CRN/2021/0089).  Subjects granted permission for image research and publication. 

\bibliographystyle{IEEEtran}

\bigskip

\medskip

\section*{Author Affiliations and Contact Information}

\footnotesize{\ \\
John Daugman and Cathryn Downing are with the Department of Computer Science and Technology,
\ University of Cambridge, \ United Kingdom.  \\
\hspace*{0.1in} e-mail:  \{John.Daugman, Cathryn.Downing\}\url{@CL.cam.ac.uk}\\
\ \\
Oluwatobi Noah Akande is with the Computer Science Department, Baze University, Abuja, Nigeria.
e-mail: \url{oluwatobi.akande@bazeuniversity.edu.ng}\\
\ \\
Oluwakemi Christiana Abikoye is with the Computer Science Department, University of Ilorin, Kwara State, Nigeria.
e-mail: \url{abikoye.o@unilorin.edu.ng} }

\begin{IEEEbiography}[{\includegraphics[width=1in,height=1.25in,clip,
keepaspectratio]{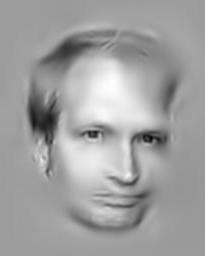}}]{John Daugman}
received his degrees at Harvard University and then
taught at Harvard before coming to Cambridge University, where he is
Professor of Computer Vision and Pattern Recognition.  He has held
the Johann Bernoulli Chair of Mathematics and Informatics at the
University of Groningen, and the Toshiba Endowed Chair at the Tokyo
Institute of Technology.  His areas of research and teaching at
Cambridge include computer vision, information theory, neural computing,
and statistical pattern recognition.  Awards for his work in science
and technology include the Information Technology Award and Medal of
the British Computer Society, the ``Time 100" Innovators Award, and the
OBE, Order of the British Empire.  He has been elected a Fellow of:
the Royal Academy of Engineering; the US National Academy of Inventors;
the Institute of Mathematics and its Applications; the British Computer
Society; and he has been inducted into the US National Inventors Hall of
Fame.  He is the founder and benefactor of the Cambridge Chrysalis Trust.
Here he is represented by a sparse sum of 2D Gabor wavelets in six orientations
and five frequencies.
\end{IEEEbiography}

\begin{IEEEbiography}[{\includegraphics[width=1in,height=1.25in,clip,
keepaspectratio]{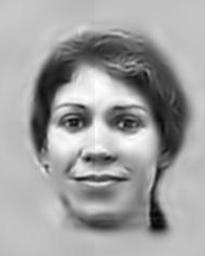}}]{Cathryn Downing}
received her B.A.\ and her Ph.D.\ degrees at Yale University and Stanford University, respectively.
Following post-doctoral appointments at New York University and at Harvard University,
she became a Research Associate at Cambridge University.  She is a Trustee of the Cambridge
Chrysalis Trust.  She and Daugman have published together on several subjects related
to the present article including:  demodulation codes; image quality metrics and predictive
norms; radial correlations within iris patterns; the universality of the IrisCode impostors'
distribution; and the effect of severe image compression on iris recognition.  She is
represented here by a linear combination of just several hundred local 2D Gabor wavelets
having six orientations, two quadrature phases, and five frequencies, each an octave
apart, thereby spanning four octaves in the same discrete self-similar log-polar system
of image representation as is used here for her co-authors' synthesized photographs.
\end{IEEEbiography}

\begin{IEEEbiography}[{\includegraphics[width=1in,height=1.25in,clip,
keepaspectratio]{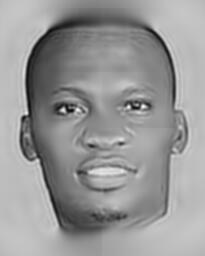}}]{Oluwatobi Akande}
is a Senior Lecturer in the Department of Computer Science, Baze University,
Abuja. He obtained a Bachelor of Technology and Masters of Technology degrees in Computer Science
from the Department of Computer Science and Engineering, Ladoke Akintola University of Technology,
Ogbomoso, Nigeria in 2012 and 2015 respectively. He also obtained a Doctor of Philosophy degree in
Computer Science from the Department of Computer Science, University of Ilorin, Nigeria in 2021.
He has passion for problem oriented researches which employ the knowledge of computing he has acquired
over the years to solve his immediate societal problems.  However, his research interest include
Biometrics, Data and Information Security, Medical Image Analysis and Machine learning. The output
of his research has been published in more than 60 journals and international conferences. He is an
active member of professional bodies such as: IAENG Computer Science, Nigeria Computer Society (NCS),
Academics in IT Professions (AITP) and Computer Professionals Registration Council of Nigeria (CPN). 
His photograph is reconstructed here using the same set of 2D Gabor wavelets in a linear combination
of six orientations, two quadrature phases, and five frequencies, each an octave apart, 
as in the synthesized images of his co-authors. 
\end{IEEEbiography}

\begin{IEEEbiography}[{\includegraphics[width=1in,height=1.25in,clip,
keepaspectratio]{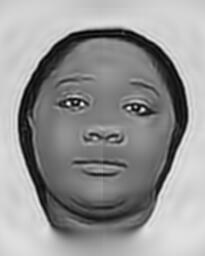}}]{Oluwakemi Abikoye}
is an Associate Professor at the Department of Computer Science,
Faculty of Communication and Information Sciences, University of Ilorin, Ilorin, Nigeria.
She received her B.Sc degree in Computer Science from the University of Ilorin in 2001, M.Sc degree
in Computer Science from University of Ibadan, Ibadan in 2006 and Ph.D.\ degree in Computer Science
also from the University of Ilorin in 2013. She began her academic career at University of Ilorin,
Department of Computer Science in 2004 as a Graduate Assistant and rose through the ranks.  Oluwakemi
is known for her innovative work in Computer/Communication Network Security.  She is the author or
coauthor of more than 90 papers in International, National and Local refereed Journals and Conference
contributions. Her research interests include Cryptography, Computer and Communication Network (Cyber)
Security, Biometrics, Human Computer Interaction and Text and Data Mining. She is also a member of the
following learned / professional societies such as Nigeria Computer Society (NCS), Science Association
of Nigeria (SAN), IEEE and IEEE Computer Society, Computer Professionals Registration Council of Nigeria
(CPN), Nigerian Women in Information Technology (NIWIIT), and Academia in Information Technology
Profession (AITP).  She is currently the Deputy Director (Research), Centre for Research, Development
and In-house Training (CREDIT).  She is represented in this image by a linear combination of the
same set of localised, multi-scale, 2D Gabor wavelets as her co-authors.
\end{IEEEbiography}

\end{document}